%% file: main_preprint.tex
\documentclass[10pt]{article}
\usepackage[preprint]{tmlr}
\input{math_commands.tex}

\input{preamble_common.tex}

\title{When Does Adaptive Halting Win? Selection Rules Matter in Looped Language Models}
\author{\name Mahesh Reddy Pagadala \email \texttt{mahesh15116@gmail.com} \\
      \addr Aston University, Birmingham, UK}
\newcommand{\codeavailability}{\url{https://github.com/lonewolf15116/lookahead-halting}}
\begin{document}
\maketitle
\input{body.tex}
\bibliography{refs}
\bibliographystyle{tmlr}
\end{document}

%% file: math_commands.tex
\usepackage{amsmath,amsfonts,bm}

\def\eqref#1{equation~\ref{#1}}

\def\1{\bm{1}}

\DeclareMathAlphabet{\mathsfit}{\encodingdefault}{\sfdefault}{m}{sl}
\SetMathAlphabet{\mathsfit}{bold}{\encodingdefault}{\sfdefault}{bx}{n}



%% file: preamble_common.tex
\usepackage[T1]{fontenc}
\usepackage[utf8]{inputenc}
\usepackage{booktabs}
\usepackage{amsmath,amssymb}
\usepackage{graphicx}
\usepackage{hyperref}
\usepackage{url}
\usepackage{xurl}
\DeclareUnicodeCharacter{03C0}{\ensuremath{\pi}}
\DeclareUnicodeCharacter{03C4}{\ensuremath{\tau}}
\DeclareUnicodeCharacter{03C1}{\ensuremath{\rho}}
\DeclareUnicodeCharacter{03B4}{\ensuremath{\delta}}
\DeclareUnicodeCharacter{0394}{\ensuremath{\Delta}}
\DeclareUnicodeCharacter{03B5}{\ensuremath{\varepsilon}}
\DeclareUnicodeCharacter{03B3}{\ensuremath{\gamma}}
\DeclareUnicodeCharacter{2264}{\ensuremath{\le}}
\DeclareUnicodeCharacter{2265}{\ensuremath{\ge}}
\DeclareUnicodeCharacter{00D7}{\ensuremath{\times}}
\DeclareUnicodeCharacter{2212}{\ensuremath{-}}
\DeclareUnicodeCharacter{2026}{\ensuremath{\ldots}}
\providecommand{\tightlist}{\setlength{\itemsep}{0pt}\setlength{\parskip}{0pt}}

%% file: body.tex
\providecommand{\tightlist}{\setlength{\itemsep}{0pt}\setlength{\parskip}{0pt}}
\emph{Evidence labels used throughout: {[}C{]} confirmatory, {[}S{]} pre-specified secondary, {[}P{]} post hoc emphasis or interpretation, {[}E{]} exploratory.}

\begin{abstract}
Looped language models can stop iterating early, and recent work shows that post-hoc readouts of the loop trajectory often rival the
learned exit gate of the looped model Ouro. Such comparisons reduce each method to one operating point chosen by a selection rule. We ask
whether that rule decides the outcome. On a frozen Ouro-1.4B with four loops, we compared a trained stability probe (π1-x) with Ouro's
native gate, adapted to a per-question stopping decision, in pre-registered studies; the confirmatory study was designed after an exploratory finding on development data and evaluated once on a sealed test set of 8,517 questions. In the pre-registered confirmatory
comparison, which selected operating points with a bootstrap-screened 1 pp accuracy-loss rule, the probe saved more of the four-loop budget
than the native gate (+0.053 gross macro saving over three benchmarks, 95\% CI {[}+0.049, +0.057{]}), mostly from CommonsenseQA, where the probe's test upper bound on accuracy loss exceeded 1 pp. Under a point-only rule, a pre-specified
secondary analysis on the same test set reversed the preference (−0.022 {[}−0.026, −0.018{]}), with operating points that exceeded the budget.
In pre-specified secondary comparisons, the probe showed no demonstrated advantage over an untrained KL divergence between successive
loops (+0.00125 {[}−0.0041, +0.0065{]}); we emphasise this result post hoc and claim no equivalence. The probe also broke its budget on one
of two out-of-distribution benchmarks (3.26 pp loss on Social IQa). Savings are estimated from stored full-depth traces, not measured
inference time, and our screen carries no formal risk-control guarantee. We conclude that halting methods should be compared as signals
together with selection procedures, with strong untrained baselines and out-of-distribution checks.
\end{abstract}

\section{Introduction}

Looped, or recurrent-depth, language models apply one block of layers several times. The number of loops therefore becomes a choice made at
inference time \citep{Zhu2025,Geiping2025,Bae2025}. If a model could stop looping once further loops no longer change its answer, it could
spend less compute on easy inputs. Ouro \citep{Zhu2025} ships a learned exit gate for this purpose. Recent work has shown that simple post-hoc
readouts of the loop trajectory, such as confidence, entropy, and divergence or displacement between loops, often match or beat that gate at
validation-selected operating points \citep{Popescu2026}. Popescu et al.~conclude that adaptive depth is ``a trajectory--readout problem, not a
gate-learning problem''.

Comparisons of this kind have one feature in common. Every method is reduced to a single operating point, chosen on held-out data by some
selection rule, such as a target accuracy or a budget on accuracy loss, and the methods are then compared at those points. The rule is usually
treated as a detail of the protocol. This paper asks whether it is one. We study three questions on a frozen Ouro-1.4B:

\begin{itemize}
\tightlist
\item
  \textbf{RQ1.} Does the selection rule change which halting method is preferred?
\item
  \textbf{RQ2.} Does a trained probe added on top of the frozen model save more than (a) Ouro's own trained gate and (b) a training-free
  divergence between successive loops?
\item
  \textbf{RQ3.} Do policies selected on development data keep their accuracy budget on benchmarks that were not used for development?
\end{itemize}

We answer these questions with a sequence of pre-registered studies. A first study (G3a) selected operating points with a point-only rule:
accuracy loss of at most 1 percentage point (pp) relative to the full four loops. Under that rule, a trained probe on cheap trajectory features
(π1-x) saved less than Ouro's gate, and the study stopped {[}E{]}. A second, confirmatory study (C1) was pre-registered after an exploratory analysis (E1) had found, on development data, that π1-x led under a stricter rule; C1 tested whether that finding held on new data.
It compared the same two signals under that bootstrap-screened rule, which also required the bootstrap 95\% upper bound of the loss to be
at most 1 pp, and it was evaluated once on a sealed test set of 8,517 questions.

\textbf{Findings.}

\begin{enumerate}
\def\labelenumi{\arabic{enumi}.}
\tightlist
\item
  Under the bootstrap-screened rule, π1-x saved more of the four-loop budget than Ouro's gate: +0.053 in gross macro saving, 95\% CI
  {[}+0.049, +0.057{]}, over OpenBookQA, CommonsenseQA and HellaSwag {[}C{]}. Under the point-only rule, applied to the same test set, the
  preference reversed: −0.022 {[}−0.026, −0.018{]} {[}S{]}. The two rules select different operating points, so what reverses is the preference that each rule produces, not the
  order of fixed policies. The point-only test operating points also exceeded the 1 pp budget on most benchmarks.
\item
  π1-x showed no demonstrated advantage over an untrained KL divergence between successive loops' answer distributions: +0.00125
  {[}−0.0041, +0.0065{]}. This comparison was pre-specified {[}S{]}. The emphasis we give it was decided after seeing the test results {[}P{]}, and the
  result is non-superiority, not equivalence.
\item
  On one of two out-of-distribution benchmarks (Social IQa), π1-x broke its accuracy budget by halting early on most questions: loss 3.26
  pp, upper bound 4.72 pp {[}S{]}. Ouro's gate stayed within the budget.
\item
  Under the primary bootstrap-screened rule, all nine benchmark-by-policy point losses on the test set were within 1 pp, but four of the
  nine corresponding upper bounds exceeded it {[}S{]}. This is compatible with several causes that we cannot separate: small samples (one
  exceedance is on HellaSwag, n = 750, for every method), the heuristic nature of our screen, and the difference between development and
  test conditions {[}P{]}.
\end{enumerate}

\textbf{Contributions.} We do not propose a new halting signal, and we do not claim that post-hoc readouts beating Ouro's gate is new; that was
shown by Popescu et al.~Our contributions are empirical and methodological:

\begin{itemize}
\tightlist
\item
  A pre-registered, sealed-test evaluation of halting signals on a frozen looped LM, in which two pre-specified operating-point selection rules led to opposite preferences (RQ1). That the preferred method can vary with the
  operating point is already visible in the crossing curves of Popescu et al.; we test it under concrete, pre-specified rules and report whether the selected points meet the accuracy budget.
\item
  Evidence that a trained probe adds no demonstrated value over a training-free convergence readout under that rule (RQ2), and an
  out-of-distribution budget failure for the trained probe (RQ3).
\item
  Six recommendations for evaluating halting methods (Section 7.4). The central one is to treat each comparison as a comparison of signal
  plus selection procedure.
\end{itemize}

\textbf{Scope.} All results are for Ouro-1.4B with four loops, multiple-choice scoring, and one halting decision per question. The policies were
evaluated from stored full-depth traces, not run with truncated computation; a truncated-execution check on 600 development questions agreed with this evaluation on every exit loop, prediction and correctness (Section 3.6). Our loop savings are therefore estimates, not measured time,
and they are not measurements of Ouro's released inference path, which selects a readout per token after computing all loops (Section 3.2).

\section{Related work}

\textbf{Learned halting for recurrent computation.}

\begin{itemize}
\tightlist
\item
  ACT \citep{Graves2016} adds a sigmoid halting unit to an RNN. It halts when the cumulative halting probability reaches 1−ε and trains with a
  ponder cost.
\item
  PonderNet \citep{Banino2021} replaces this with a learned conditional halting distribution and a geometric prior.
\item
  Ouro's exit gate \citep{Zhu2025} belongs to the same family. A linear head on each loop's state gives a per-loop exit probability, and Q-exit
  stops at the first loop whose cumulative exit probability reaches a threshold q. The gate is trained jointly with the model and then
  refined on the loss improvement from an extra loop.
\item
  Ouro's own analysis on MMLU found the trained gate best at every budget. A hidden-state difference threshold came within one to two
  points of it \citep[section 5.4.1]{Zhu2025}.
\item
  Joint training of depth controllers is also studied in Mixture-of-Recursions \citep{Bae2025} (per-token routers over recursion depth), in
  AdaPonderLM \citep{Song2026} (per-iteration gates with a ponder penalty) and in Think-at-Hard \citep{Fu2025} (a per-token iteration decider). We train
  nothing in the backbone or the gate; our probe sits on top of a frozen model.
\end{itemize}

\textbf{Post-hoc readouts on Ouro.} The closest work is \citet{Popescu2026}:

\begin{itemize}
\tightlist
\item
  They evaluated the released Ouro-1.4B and 2.6B on six benchmarks, including the three primary benchmarks we use.
\item
  They compared the pretrained gate with post-hoc readouts: entropy, top-1 confidence, margin, predictive KL, logit change, and hidden-state
  displacement and cosine change.
\item
  They chose thresholds on validation data and evaluated on held-out data.
\item
  They found the gate ``competitive but not uniformly Pareto-optimal''. Their full per-benchmark curves interleave, so the preferred method
  visibly varies with the operating point in their data, though they report one validation-selected point per method and do not state how it was chosen.
\item
  Our readout definitions follow theirs \citep[appendix B.6]{Popescu2026}. Our threshold grid has a similar size but a different construction:

  \begin{itemize}
  \tightlist
  \item
    ours: 50 quantiles over the full range plus a never-exit point, inherited from earlier stages of this project;
  \item
    theirs: 51 quantiles between the 0.5th and 99.5th percentiles \citep[appendix B.5]{Popescu2026}.
  \end{itemize}
\item
  Our work differs from theirs in four respects:

  \begin{enumerate}
  \def\labelenumi{\arabic{enumi}.}
  \tightlist
  \item
    a trained probe on the frozen trajectory;
  \item
    per-benchmark accuracy-loss budgets with a fixed-depth baseline;
  \item
    a comparison of two selection rules;
  \item
    a pre-registered evaluation, run once on a sealed test set.
  \end{enumerate}
\end{itemize}

\textbf{Convergence and stability exits.}

\begin{itemize}
\tightlist
\item
  \citet{Geiping2025} stop a recurrent-depth model once the KL divergence between successive steps falls below 5\(\times 10^{-4}\), per
  generated token. Our r-kl applies the same kind of signal to the answer-option distribution, once per question, with thresholds selected
  per benchmark.
\item
  \citet{Pappone2025} propose a second-order (``acceleration'') criterion on the step size and report that it beats the KL exit in a
  small recurrent-depth model.
\item
  PABEE \citep{Zhou2020} stops a fixed-depth encoder once consecutive layer classifiers agree for a set number of layers. It is the layer-wise
  ancestor of the stability target we train π1-x on.
\item
  CALM \citep{Schuster2022} exits decoder layers per token using softmax confidence, hidden-state saturation or a trained classifier.
\end{itemize}

\textbf{Risk-controlled early exit.} Choosing exit thresholds with statistical guarantees is established:

\begin{itemize}
\tightlist
\item
  CALM calibrates its thresholds with Learn then Test \citep{Schuster2022,Angelopoulos2025LTT}.
\item
  Fast yet Safe \citep{Jazbec2024} selects early-exit thresholds with a UCB bound. It controls the performance gap to the full model with
  probability 1−δ under i.i.d. calibration data, and leaves distribution shift to future work.
\item
  Conformal risk control \citep{Angelopoulos2024CRC} controls the expected value of a monotone loss.
\item
  \citet{Kubaty2026} show that calibration alone is insufficient to measure the quality of early-exit classifiers.
\end{itemize}

Our bootstrap-screened rule is none of these procedures. Its bound is asymptotic, uncorrected for the number of candidate thresholds, and
computed on the data that selected the candidate. Our results therefore inherit none of their guarantees (Section 3.4). Part of our
contribution is to show what happens on a sealed test set when such a heuristic screen is used instead.

\textbf{Monitoring and engineering evaluations of Ouro's exit.}

\begin{itemize}
\tightlist
\item
  \citet{Olszowiec2026} find that in most of eight looped models the large majority of generated tokens settle on their final
  top-1 prediction at the first iteration, and that Ouro's gate fires several iterations after the answer is fixed.
\item
  An engineering evaluation of Ouro-1.4B's native exit on GSM8K reported a 7.35 pp accuracy loss at about 0.55 of the loops for one gate
  setting \citep{vllmrlt56}. This is a GitHub issue, not peer-reviewed.
\end{itemize}

Both suggest that the native gate leaves savings unused in generation. Our setting, a single per-question decision in multiple-choice
scoring, is different and does not test that claim.

\textbf{Early exit of reasoning chains.} SpecExit \citep{Yang2026}, NEAT \citep{Liu2026} and hidden-state correctness probes for chain-of-thought \citep{Zhang2025}
shorten the generated reasoning text, not the latent depth. We cite them as adjacent work and do not compare numbers.

\section{Setting: model, halting signals and selection rules}

\subsection{Model, scoring and decision unit}

We use Ouro-1.4B \citep{Zhu2025}, frozen, in bf16, with its trained depth of 4 loops. No weight is changed. Every benchmark is multiple choice. Each answer option is scored by the summed log-likelihood of its continuation tokens, using packed scoring: the context and all options form one sequence, each option token may attend only to the context and to earlier tokens of its own option, and one forward pass per loop yields all option scores. The answer is the option with the highest score, using the length-normalised log-likelihood (the \texttt{acc\_norm} metric) for OpenBookQA, HellaSwag and PIQA, and the raw log-likelihood (\texttt{acc}) for CommonsenseQA and Social IQa. OpenBookQA is evaluated 0-shot, CommonsenseQA and HellaSwag 10-shot, and the two out-of-distribution benchmarks 0-shot.

A halting policy makes one decision per question. After loop \(t \in \{1,2,3\}\) it reads Ouro's exit-gate output, the hidden state and the full-vocabulary next-token entropy at the last context token, together with the option-level scores at loop \(t\), and either stops or continues. Loop 4 is the reference depth. For a question that exits at loop \(e \in \{1,2,3,4\}\) the answer is the loop-\(e\) answer, and the accuracy loss of a policy on a benchmark is \(\text{acc}(\text{loop }4) - \text{acc}(\text{policy})\), in percentage points.

\subsection{Ouro's native gate and what its released implementation does}

Ouro attaches an exit gate to each loop: \(\lambda_t = \sigma(\text{Linear}(h_t))\). The exit distribution is \(p_t = \lambda_t \prod_{j<t}(1-\lambda_j)\), with the last loop taking the remaining mass. Its cumulative sum is the CDF, and a Q-exit stops at the first loop whose CDF reaches a threshold \(q\) \citep{Zhu2025}.

The released implementation (\texttt{modeling\_ouro.py}, sha256 c5c68fbb\ldots, in the pinned model revision) does not skip loops. Its forward pass runs all \texttt{total\_ut\_steps} loops, applies the gate (a \texttt{Linear(hidden,\ 1)}) at every token position, and, when an exit threshold is given, chooses for each token position the first loop whose cumulative exit probability reaches the threshold and then reads out that loop's hidden state through the language-model head. The released Q-exit therefore operates per token position and selects a readout; it saves no compute. The paper's text states the goal of exiting per token and per example but does not fix the inference unit.

Our use is an adaptation. We evaluate the gate's CDF at a single position, the last context token, which gives one decision per question, and we count the loops after the exit as saved. Two consequences follow, and we state them wherever savings are reported. First, our native-gate baseline uses Ouro's trained gate as a halting statistic inside our own protocol; it is not Ouro's released inference procedure. Second, our savings are properties of this adaptation. They are not measurements of the released inference.

\subsection{Halting signals}

All policies share one form. Each has a statistic \(s_t\) for loops \(t = 1,2,3\), oriented so that larger means ``safer to stop'', a threshold \(\tau_b\) and a depth cap \(m_b \in \{1,2,3,4\}\), both chosen per benchmark \(b\):

\[e = \min\{\, t < m_b : s_t \ge \tau_b \,\} \quad \text{if this set is non-empty, and } e = m_b \text{ otherwise.}\]

A statistic that is undefined at loop 1 never fires there. Setting \(\tau_b = +\infty\) gives fixed depth \(m_b\). The suffix ``-pb'' marks per-benchmark selection of \((\tau_b, m_b)\).

\textbf{Native gate (native-pb).} \(s_t = \text{CDF}_t = 1 - \prod_{j \le t}(1-\lambda_j)\), computed from Ouro's own gate logits at the last context token. It has no fitted component.

\textbf{Learned probe (π1-x).} A per-loop L2-regularised logistic regression, one model for each of loops 1, 2 and 3, with regularisation constant \(C = 1.0\) (inherited from the Gate 2 configuration, not re-tuned) and standardised inputs. The input has nine columns:

\begin{itemize}
\tightlist
\item
  conf: \(\max_k p_t(k)\), where \(p_t\) is the softmax of the option scores;
\item
  ent: the entropy of \(p_t\);
\item
  vent: the entropy of the full-vocabulary next-token distribution at the last context token;
\item
  cdf: the native gate's CDF at loop \(t\);
\item
  n\_options: the number of answer options;
\item
  log\_n\_tokens: the logarithm of the packed sequence length (context plus all options);
\item
  three loop indicators, which are constant within each per-loop model and so carry no information there.
\end{itemize}

The target is the stability label \(S_t\): it equals 1 if the loop-\(t\) predicted option is the same as the predicted option at every later loop, so that the answer never changes again. The label is an agreement criterion in the spirit of patience-based exits over layers \citep{Zhou2020}, used here as a training label for a probe over loops rather than as a patience counter. The score is \(s_t = \hat{P}(S_t = 1)\). The probes were fitted on development data from all six development benchmarks. For selection, \(s_t\) is the five-fold cross-fitted out-of-fold score over all development data; for TEST, the final models were refitted on all development data and saved.

\textbf{Untrained readouts (r-*).} Seven rules computed from the stored per-loop option scores \(z_t\) (the scores whose softmax is \(p_t\)) or hidden states \(h_t\) at the last context token, following \citet{Popescu2026} and Ouro's hidden-state rule \citep{Zhu2025}:

\begin{center}\small
\begin{tabular}{@{}lll@{}}
\toprule
Rule & Statistic \(s_t\) & Loops with \(s_t\) defined \\
\midrule
r-conf & \(\max_k p_t(k)\) & 1--3 \\
r-margin & \(z_t^{(1)} - z_t^{(2)}\), the gap between the top two option scores & 1--3 \\
r-ent & \(-H(p_t)\) & 1--3 \\
r-kl & \(-\mathrm{KL}(p_t \,\|\, p_{t-1})\) & 2--3 \\
r-dlogit & \(-\lVert z_t - z_{t-1} \rVert_2\) & 2--3 \\
r-hl2 & \(-\lVert h_t - h_{t-1} \rVert_2\) & 2--3 \\
r-hcos & \(\cos(h_t, h_{t-1})\) & 2--3 \\
\bottomrule
\end{tabular}
\end{center}

\textbf{Fixed depth (B\_pb).} \(\tau_b = +\infty\), with the depth \(m_b\) chosen per benchmark by the same rule as the adaptive policies. It is a reference, not the primary comparator.

\subsection{\texorpdfstring{Selection of \((\tau_b, m_b)\)}{Selection of (\textbackslash tau\_b, m\_b)}}

Selection uses development data only, separately for each benchmark. The candidate set for a statistic is the product of a threshold grid and \(m_b \in \{1,2,3,4\}\). The grid holds up to 50 distinct quantile values of the statistic, pooled over the usable questions of all development benchmarks and the loops where it is defined, plus \(+\infty\). This gives up to 51 thresholds and 204 candidates per benchmark and policy. Two rules decide which candidates qualify.

\begin{itemize}
\tightlist
\item
  \textbf{Point-only rule} (the G3a rule). A candidate qualifies if its point accuracy loss on the selection data is at most 1 pp, with a numerical tolerance of 1e-12.
\item
  \textbf{Bootstrap-screened rule} (called the ``combined risk rule'' in the C1 protocol, and implemented as \texttt{select\_risk} in the E1 code). A candidate qualifies if its point loss is at most 1 pp and the one-sided 95\% bootstrap upper bound of its loss on the selection data is also at most 1 pp.~The bound is the 95th percentile of the loss over 2,000 resamples of the selection questions of that benchmark, with a fixed random seed per benchmark and the same resampled draws for every policy.
\end{itemize}

Among qualifying candidates the rule takes the maximum gross saving; ties within 1e-12 are broken by smaller point loss, then by preferring fixed depth (\(\tau = +\infty\)), then by larger \(\tau\), then by smaller \(m\). Fixed depth 4 has zero loss and always qualifies.

The bootstrap-screened rule is a heuristic screen. It carries no formal guarantee of risk control, and we do not describe it as risk control. Prior work selects early-exit thresholds with finite-sample statements: Fast yet Safe uses a high-probability upper confidence bound on a performance gap under i.i.d. calibration data \citep{Jazbec2024}; Learn then Test frames threshold selection as multiple testing with family-wise error control \citep{Angelopoulos2025LTT}; conformal risk control bounds the expected value of a monotone loss \citep{Angelopoulos2024CRC}. Our rule differs in four ways: the bootstrap is asymptotic, not a finite-sample bound; there is no multiplicity correction across the candidates; selection and bound use the same data, so the selected candidate's bound is optimistic; and no \((\alpha, \delta)\) statement is made. The bootstrap also treats the cross-fitted scores as fixed, so it ignores the variability from fitting the probe. The held-out record is consistent with a screen that does not control held-out risk: in the exploratory study E1 the screened rule still left 3 of 15 held-out benchmark-by-fold cells above 1 pp for both native-pb and π1-x {[}E{]}.

\subsection{Gross and net saving}

For a question exiting at loop \(e\), the gross saving is \(1 - e/4\). A benchmark's gross saving is the mean over its questions, and the macro saving is the unweighted mean over the three primary benchmarks.

The net saving charges π1-x for the probe's extra work. The native gate's CDF comes from the model's exit head, which is shared and counted for no policy, so native-pb and fixed depth have \(\rho = 0\). For π1-x two overheads are charged in units of one loop's time: \(\rho_{\text{score}}\), the cost of scoring the options at a loop where the controller evaluates but does not stop, and \(\rho_{\text{rest}}\), the cost of the vocabulary entropy, the host transfer, the nine features and the probe, charged at every evaluation. Scoring at the halting loop is not charged, because it is the scoring that produces the answer and every controller needs it. For a question that halts by its score at loop \(e < m_b\), the evaluations number \(e\) and the extra scorings number \(e-1\); for a question that reaches the cap, both number \(m_b - 1\). The net saving of a question is

\[1 - \frac{e + \rho_{\text{score}}\, n_{\text{score}} + \rho_{\text{rest}}\, n_{\text{eval}}}{4}.\]

The overheads \(\rho_{\text{score}}\) and \(\rho_{\text{rest}}\) were measured once on the GPU at freeze time, on 200 development questions per primary benchmark at batch size 1, using extraction's own code path (Table 1). The net saving is an estimated deployment cost under this accounting boundary. It is not a wall-clock measurement.

\begin{table}[t]
\centering\small
\caption{Freeze-time overhead of the probe, as a fraction of one loop's time. Measured on one GPU, batch size 1, 200 development questions per benchmark.}
\resizebox{\ifdim\width>\linewidth\linewidth\else\width\fi}{!}{%
\begin{tabular}{@{}lllll@{}}
\toprule
Benchmark & \(\rho_{\text{score}}\) & \(\rho_{\text{rest}}\) & \(\rho_{\text{total}}\) & One loop (ms) \\
\midrule
OpenBookQA & 0.036 & 0.014 & 0.050 & 20.7 \\
CommonsenseQA & 0.010 & 0.004 & 0.014 & 79.9 \\
HellaSwag & 0.013 & 0.003 & 0.015 & 148.8 \\
\bottomrule
\end{tabular}
}
\end{table}

\subsection{Retrospective evaluation}

All stopping policies in this paper were evaluated retrospectively, offline, from full-depth per-loop traces. They were not executed with truncated computation. The extraction stage ran every question through all four loops once and stored, for each loop, the option scores and distributions, the prediction and its correctness, the gate logit and CDF, the vocabulary entropy and the hidden state at the last context token. The evaluation code then applies a policy to those stored arrays: it finds the exit loop from the stored statistics and reads the correctness at that loop from the stored correctness array. No halting policy was run through the model.

Accuracy and exit-loop outcomes obtained this way would be exact for a truncated run, provided that every quantity a policy reads at loop \(t\) depends only on loops up to \(t\). Our reading of the code is that this condition holds, and a diagnostic on development data supports it {[}E{]}.

\textbf{Truncated-execution check (D-t).} We ran the three frozen policies (π1-x, native-pb, r-kl) with real early stopping on 600 development questions, 200 per primary benchmark, and compared the results with the stored-trace evaluation of the same questions. The test set was not used.

\begin{itemize}
\tightlist
\item
  Exit loop, predicted option and correctness agreed in all 1,800 policy-question pairs. The differences in accuracy and simulated saving were therefore zero.
\item
  A model-free check that masks later loops in the stored arrays found no dependence of any policy statistic on later loops (0 differences in 5,594 checks). This supports the causality condition for the tested statistics and inputs.
\item
  The pre-declared bit-for-bit comparison of the statistics failed on HellaSwag for the two policies that read option scores. Separate GPU executions of the same inputs differed by up to 1.6\(\times 10^{-7}\) in the statistic and 9.2\(\times 10^{-5}\) in option log-likelihoods. Values read at the last context token were bit-identical. A follow-up diagnostic (D-t2) supports GPU floating-point nondeterminism as a cause {[}E{]}. We ran 20 HellaSwag questions 5 times each at full depth: under default settings, 16 of 20 questions showed bit differences across repeats, while with deterministic algorithms enabled (\texttt{torch.use\_deterministic\_algorithms(True)}, \texttt{CUBLAS\_WORKSPACE\_CONFIG=:4096:8}) none did. This establishes reproducibility only within that deterministic configuration. It does not turn the original bitwise comparison into a pass. These differences changed no observed decision.
\end{itemize}

The check therefore does not establish bitwise reproducibility. It also does not establish agreement on the test set, or any latency or runtime saving. We also did not measure whether halting decisions change under other numerical paths (Section 7.5). The loop runner repeats the layer stack on a single packed sequence, the input to loop \(t\) is the normalised output of loop \(t-1\), attention at loop \(t\) uses keys and values computed from that loop's own input (the runner keeps no cache across loops), and the gate CDF at loop \(t\) is a function of the gate logits at loops up to \(t\). The statistics used by the policies at loop \(t\) (option scores, \(p_t\), entropies, probe features, the divergence from \(p_{t-1}\), hidden-state differences) are functions of loops up to \(t\). The stability label \(S_t\) looks at later loops, but only as a training label on development data; it is never an input at decision time.

Compute savings are of a different kind. They are simulated from exit-loop counts, \(1 - e/4\), and the probe overhead \(\rho\) was measured separately at freeze time, in isolation, on development questions (Table 1). They are therefore an estimate of the savings a truncated implementation could realise, under the assumption that a skipped loop costs what a measured loop costs. The reported policy savings are retrospective estimates rather than observed end-to-end savings. Separately, D-t collected descriptive batch-size-1 timings on 600 development questions; these timings are not a held-out latency evaluation and support no general runtime-speedup claim.

\section{Experimental design and pre-registration}

\subsection{Data}

Development data are the Phase 1 training questions (17,615) and the Phase 3 development allocation, DEV+ (2,635), together 20,250 questions across six benchmarks: OpenBookQA, CommonsenseQA, HellaSwag, ARC-Challenge, ARC-Easy and MMLU. They were used to fit the probes, select every \((\tau_b, m_b)\), measure \(\rho\) and freeze the study. A further 5,870 Phase 1 validation questions were exposed earlier and were used for a report-only estimate in G3a; the C1 protocol excludes them from every choice (the protocol's reviewer decision D-a, ``development data only'').

The TEST set is a sealed allocation of 8,517 questions, disjoint from the development data by a salted hash and fixed in the manifest (sha256 b2df1930\ldots). The primary in-distribution benchmarks are OpenBookQA (3,000), CommonsenseQA (1,000) and HellaSwag (750), 4,750 questions in all. The out-of-distribution benchmarks, PIQA (1,837) and Social IQa (1,930), are reported separately and never pooled. ARC and MMLU have no fresh rows and are not part of any claim. Every allocated row was removed from the few-shot exemplar pool, and the out-of-distribution training splits were never used.

\begin{table}[t]
\centering\small
\caption{Data split.}
\begin{tabular}{@{}lll@{}}
\toprule
Set & Questions & Role \\
\midrule
Development (Phase 1 train + DEV+) & 17,615 + 2,635 = 20,250 & fitting, selection, freeze \\
TEST, OpenBookQA & 3,000 & primary \\
TEST, CommonsenseQA & 1,000 & primary \\
TEST, HellaSwag & 750 & primary \\
TEST, PIQA & 1,837 & out of distribution \\
TEST, Social IQa & 1,930 & out of distribution \\
TEST, total & 8,517 & one-time confirmatory run \\
\bottomrule
\end{tabular}
\end{table}

\subsection{Sequence of studies}

The studies were run in a fixed order, and each later one was written after the earlier one was closed.

\begin{enumerate}
\def\labelenumi{\arabic{enumi}.}
\tightlist
\item
  \textbf{G3a.} The pre-registered development gate of Phase 3 (plan v1.1) compared π1-x and a hidden-state variant with the strongest baseline under the point-only rule, using nested cross-validation on development data. It stopped at rule D2, because neither candidate improved on native-pb by the required 0.02. The decision was computed by the frozen code and confirmed by the reviewer.
\item
  \textbf{E1 {[}E{]}.} An exploratory analysis on development data asked why the native gate won. Its main finding was that the ordering depended on the selection rule: under the point-only rule π1-x trailed native-pb (−0.024), whereas under the bootstrap-screened rule applied identically to both it led (+0.053, positive in all five folds).
\item
  \textbf{C1 protocol.} E1 motivated a new, separately pre-registered study, C1, whose single confirmatory test is on the sealed TEST set. The protocol was reviewed (conditional approval with five specified decisions: development data only, δ = 0.02, gross saving primary with net saving a mandatory secondary, 10,000 resamples with seed 2026, native-pb as the sole primary comparator), then frozen (sha256 4f38b8b7\ldots, tag \texttt{c1-protocol}).
\item
  \textbf{Freeze on development data.} One script fitted π1-x, ran the selections for both policies, the fixed-depth reference and the pooled out-of-distribution selection, measured \(\rho\), and wrote \texttt{frozen.json} (65728e16\ldots) and the model file (72f15508\ldots).
\item
  \textbf{One-time guarded TEST run.} The run applied the frozen selections once to the sealed TEST set.
\end{enumerate}

The guard around the TEST run is part of the methodological rigour of the study. We do not present it as a contribution. Before reading any outcome, the runner verified the sha256 of the protocol, the frozen selections, the model file, the manifest and the TEST extraction check; verified the code files recorded at freeze; verified that the 24-file code bundle matched the approved hash (c0902af3\ldots); refused to start if the marker, the report or a failure record already existed; and checked the TEST extraction's metadata (identifier lists, counts, file checksums) without reading any prediction or label. It then created the marker \texttt{TEST\_OPENED.json} by exclusive creation, so that it could never be overwritten, and only then read the TEST extraction, restricted to the TEST directory. The report was written by exclusive creation. A second invocation after the successful run was refused at preflight, because the marker existed, and TEST was not read again.

\subsection{Frozen decision rule, quoted from the protocol}

The endpoint and the budget are quoted from \texttt{docs/C1\_PROTOCOL.md} (sha256 4f38b8b7\ldots), sections C1.3 to C1.5 (adjacent bullets are joined into paragraphs; the wording is unchanged). The hypothesis statement in C1.1 is paraphrased, not quoted, below.

\begin{quote}
\textbf{Primary.} Δ = S\_π1-x − S\_native, the paired difference in \textbf{gross} loop saving, macro-averaged with equal weight over the three primary benchmarks on TEST. \textbf{Minimum meaningful improvement δ = 0.02} (2 percentage points of loops; the same size as Phase 3's D2 screen).

\textbf{Budget, per primary benchmark.} loss\_b = acc\_b(loop 4) − acc\_b(policy) ≤ 1.00 pp, point estimate on TEST, with the Gate 1 tolerance of 1e-12. \textbf{π1-x must meet the budget on all three benchmarks.} If it does not, the result is \textbf{FAIL (budget)}, whatever Δ is. A saving advantage alone never counts as success.

\textbf{Interval.} Paired bootstrap over TEST questions, resampled within each benchmark, so each benchmark keeps its own n and the macro stays equal-weight. B = 10,000, seed \texttt{default\_rng(2026)} (\textbf{D-d}). Two-sided 95\% percentile interval {[}L, U{]} for Δ.

\textbf{Verdict (exhaustive; the first matching row applies):}

\begin{center}\small
\begin{tabular}{@{}p{0.36\linewidth}p{0.5\linewidth}@{}}
\toprule
condition & verdict \\
\midrule
π1-x loss \textgreater{} 1 pp on any primary benchmark & \textbf{FAIL (budget)} \\
Δ ≥ δ \textbf{and} L \textgreater{} 0 & \textbf{PASS}: a statistically supported positive improvement (L \textgreater{} 0) with an observed effect of at least δ = 0.02 (point estimate) \\
U \textless{} δ & \textbf{FAIL (no meaningful advantage)}: improvements of size δ or more are ruled out \\
otherwise & \textbf{INCONCLUSIVE}: Δ and {[}L, U{]} reported, no claim either way \\
\bottomrule
\end{tabular}
\end{center}
\end{quote}

\begin{quote}
\textbf{What PASS does and does not claim (reviewer clarification).} PASS establishes, at 95\% confidence, that the improvement is \textbf{positive}, and that the \textbf{observed} effect is at least 0.02. It does \textbf{not} establish with 95\% confidence that the true improvement is at least 0.02; that would need L \textgreater{} 0.02, which is not the criterion. Reports must keep the two apart. Whether L \textgreater{} 0.02 also holds is reported as a secondary fact.

\textbf{One primary test.} Everything else is secondary and never changes the verdict.
\end{quote}

In words: the comparison is between π1-x and native-pb, each selected on development data under the same bootstrap-screened rule. The protocol also fixed, before TEST was opened, the secondary analyses reported in sections 5 and 6: the same comparison under the point-only rule; π1-x against every other baseline, with native-pb as the sole primary comparator; the net saving; the out-of-distribution benchmarks, each on its own, with one task-agnostic \((\tau, m)\) per policy selected under the same rule over all six development benchmarks; and a table of held-out losses for every policy. The protocol also stated before the test that demonstrating a loss of at most 1 pp would be out of reach for CommonsenseQA (n = 1,000) and HellaSwag (n = 750), even for a policy that loses nothing: with a true loss of zero and 5\% to 10\% of questions changing correctness, the expected one-sided upper bound is 1.16 to 1.65 pp for CommonsenseQA and 1.34 to 1.90 pp for HellaSwag (planning assumptions from amendment A1.5). The expected half-width of the 95\% interval for Δ at the TEST sizes was computed at freeze from development data as 0.0043.

\subsection{Deviations and disclosures}

\begin{itemize}
\tightlist
\item
  \textbf{Phase 3 numerical diagnostic D-a (PIQA).} Before the full extraction, a float32 check compared the packed pass with a context-only pass on 100 never-used PIQA training rows and 100 Social IQa rows, against a tolerance of 1e-4. Social IQa passed (maximum difference 3.1e-5). PIQA failed (1.17e-4). The tolerance was not changed and the failure stands as recorded. A follow-up, D-a2, repeated the comparison in float64 on the CPU reference path for the 10 PIQA questions with the largest float32 difference, with a pre-fixed criterion of at most 1e-9; the maximum difference was 1.67e-13, and the result was classified numerical. The reviewer accepted a documented PIQA deviation and released the extraction. A procedural deviation within D-a2 (a one-line fix was run before reviewer confirmation) is recorded in the decision log.
\item
  \textbf{Console transcript.} After the successful TEST run, the runner was started a second time by mistake and was refused at preflight, before the marker step. Its console transcript overwrote the transcript file of the successful run, so the file holds only the refused attempt. The console output of the successful run survives as a paste in the working conversation. The run itself is fully recorded in \texttt{test\_report.json} and the marker.
\item
  \textbf{Smaller disclosures.} One test, the symlink-escape test for unsealing, was skipped on the Windows machine that ran the TEST script (it passes on Linux). The package snapshots before and after the run differ in one line, the editable install of this repository (the commit before and after the pull of the approved commit); all third-party versions are identical. \texttt{frozen.json} records code hashes but not a git commit.
\item
  \textbf{Not run.} The Phase 3 check D-b, on the sensitivity of halting decisions to the bf16 numerical path, has no recorded result (Section 7.5).
\item
  \textbf{Development-only follow-up (X1).} After the confirmatory study and an internal methodology review, we froze the protocol of one further exploratory analysis, X1 (correctness flips at matched gross saving; Section 7.3), and only then ran it on development data. It used E1's held-out folds and statistics and did not read TEST.
\end{itemize}

\section{Results: RQ1, the preferred method under two selection rules}

RQ1 asks whether the accuracy-risk selection rule changes which halting method is preferred. Each rule selects its own operating points \((\tau_b, m_b)\) for each method, so the comparison is between rule-dependent selections, not between fixed policies whose order flips.

\subsection{Primary and secondary comparisons}

\begin{table}[t]
\centering\small
\caption{π1-x minus native-pb under the two selection rules. The first two rows are development-data results; the last three are on the sealed TEST set. The G3a value to four decimals is −0.0236.}
\resizebox{\ifdim\width>\linewidth\linewidth\else\width\fi}{!}{%
\begin{tabular}{@{}llll@{}}
\toprule
Rule and data & π1-x − native-pb (gross macro saving) & 95\% CI & Evidence \\
\midrule
Point-only, G3a nested held-out folds on development data & −0.024 & per-fold range −0.044 to −0.014 & G3a STOP, {[}E{]} \\
Point-only, G3a exposed Phase 1 validation & −0.029 & none computed & {[}E{]} \\
Point-only, TEST & −0.022 & {[}−0.026, −0.018{]} & {[}S{]} \\
\textbf{Bootstrap-screened, TEST} & \textbf{+0.053} & \textbf{{[}+0.049, +0.057{]}} & \textbf{{[}C{]}, verdict PASS} \\
Bootstrap-screened, TEST, net of estimated probe overhead & +0.038 & {[}+0.033, +0.042{]} & {[}S{]}, estimated cost \\
\bottomrule
\end{tabular}
}
\end{table}

Under the bootstrap-screened rule, π1-x stayed within the point budget on every primary benchmark (point losses 0.17, 0.60 and 0.40 pp for OpenBookQA, CommonsenseQA and HellaSwag), the observed difference was at least \(\delta = 0.02\), and the lower bound was above 0, so the verdict was PASS {[}C{]}. The macro saving was 0.239 for π1-x and 0.186 for native-pb. Here the lower bound (0.049) also exceeded \(\delta\), which the protocol records as a secondary fact {[}S{]}. As the protocol states, PASS is not a 95\% claim that the true effect is at least 0.02. The realised half-width of the interval, 0.0042, was close to the 0.0043 expected at freeze {[}S{]}. The result is limited to Ouro-1.4B with 4 loops, to these three benchmarks, to gross saving and net saving as estimated here, and to this selection rule. The interval quantifies test-sampling uncertainty conditional on the frozen policies. It does not account for variability from policy fitting, operating-point selection or changes in evaluation conditions. The observed development-to-test shifts in macro saving (for example, 0.258 for π1-x in the cross-fitted E1 estimate against 0.239 on TEST; Section 6.1) illustrate the limits of interpreting this interval more broadly. The comparison is also at unmatched loss: averaged over the three benchmarks, π1-x's selected points lost 0.39 pp on TEST against 0.08 pp for native-pb (means of the Table 4 losses) {[}P{]}. Part of the saving difference therefore reflects more of the budget being used, and native-pb's screened selections left more of the budget unused. Neither rule compares the methods at equal loss.

Under the point-only rule, the order is the opposite: π1-x saves 0.022 less than native-pb on TEST {[}S{]}, replicating on sealed data the direction found in G3a and E1 on development data {[}E{]}. At G3a, π1-x improved on the comparator by −0.0236 on nested held-out folds; this was below the required +0.02, and Phase 3 stopped. In the point-only TEST selection, both policies exceed the 1 pp point budget on some benchmarks (see Table 5), so the point-only comparison on TEST compares policies that do not meet the budget. It is reported as a secondary replication of the rule dependence, not as a statement about compliant policies {[}S{]}.

\subsection{Per benchmark}

\begin{table}[t]
\centering\small
\caption{TEST results under the bootstrap-screened rule, per primary benchmark {[}S, descriptive{]}. Savings are gross. The upper bound is the one-sided 95\% paired-bootstrap bound of the loss on TEST. Bold marks bounds above 1 pp.}
\resizebox{\ifdim\width>\linewidth\linewidth\else\width\fi}{!}{%
\begin{tabular}{@{}lllllll@{}}
\toprule
Benchmark & Method & \(m_b\) & Gross saving & Loss (pp) & Upper bound (pp) & Exits at loop 1 / 2 / 3 / 4 \\
\midrule
OpenBookQA (3,000) & π1-x & 4 & 0.157 & 0.17 & 0.53 & 6 / 243 / 1,377 / 1,374 \\
OpenBookQA & native-pb & 4 & 0.136 & 0.20 & 0.67 & 0 / 26 / 1,579 / 1,395 \\
OpenBookQA & B\_pb & 4 & 0.000 & 0.00 & 0.00 & 0 / 0 / 0 / 3,000 \\
CommonsenseQA (1,000) & π1-x & 4 & 0.282 & 0.60 & \textbf{1.10} & 8 / 349 / 405 / 238 \\
CommonsenseQA & native-pb & 4 & 0.161 & −0.10 & 0.30 & 0 / 1 / 643 / 356 \\
CommonsenseQA & B\_pb & 4 & 0.000 & 0.00 & 0.00 & 0 / 0 / 0 / 1,000 \\
HellaSwag (750) & π1-x & 3 & 0.279 & 0.40 & \textbf{1.60} & 0 / 88 / 662 / 0 \\
HellaSwag & native-pb & 3 & 0.262 & 0.13 & \textbf{1.33} & 0 / 36 / 714 / 0 \\
HellaSwag & B\_pb & 3 & 0.250 & 0.53 & \textbf{1.60} & 0 / 0 / 750 / 0 \\
\bottomrule
\end{tabular}
}
\end{table}

The macro advantage is not spread evenly. Per benchmark, π1-x saved more than native-pb by 0.021 on OpenBookQA, 0.121 on CommonsenseQA and 0.017 on HellaSwag (differences of the entries in Table 4; no interval was computed for these individual differences) {[}P{]}. The difference is therefore dominated by CommonsenseQA. On OpenBookQA both adaptive policies split between loops 3 and 4. On HellaSwag, where the cap is 3, most questions exit at loop 3 (662 of 750 for π1-x, 714 of 750 for native-pb). On CommonsenseQA π1-x exits at loop 2 for 349 of 1,000 questions, where native-pb almost never does (1 of 1,000) {[}S{]}.

\subsection{Upper bounds above 1 pp on TEST}

The selection rule required each selected candidate's bootstrap upper bound to be at most 1 pp on development data. On TEST this did not hold everywhere. All nine point losses are within the 1 pp budget, but four of the nine one-sided upper bounds exceed 1 pp {[}S{]}:

\begin{itemize}
\tightlist
\item
  π1-x: CommonsenseQA 1.10 pp and HellaSwag 1.60 pp (OpenBookQA 0.53 pp is below);
\item
  native-pb: HellaSwag 1.33 pp;
\item
  B\_pb: HellaSwag 1.60 pp.
\end{itemize}

At selection time the bounds of the frozen candidates were between 0.75 and 0.98 pp for the two adaptive methods. HellaSwag, with \(n = 750\), exceeds 1 pp for every method including fixed depth, so the small sample is part of the cause, as the protocol anticipated (section 4.3). The exceedance for π1-x on CommonsenseQA is specific to π1-x {[}S{]}. CommonsenseQA is also where most of the macro advantage of π1-x arises: its per-benchmark difference from native-pb is 0.121, against 0.021 on OpenBookQA and 0.017 on HellaSwag (Section 5.2). Several causes are plausible and we cannot separate them {[}P{]}:

\begin{itemize}
\tightlist
\item
  sampling variability on the test set, which is large for HellaSwag;
\item
  selection effects of a heuristic screen whose bound is asymptotic, uncorrected for the number of candidates, and computed on the data that chose the candidate (Section 3.4);
\item
  variability that the development-time bound does not see. Selection used π1-x's out-of-fold development scores, but the policy applied on TEST used the probe refitted once on all development data. The two score distributions are therefore not identical, and the bound conditions on neither the refit nor its variability;
\item
  differences between development and test data.
\end{itemize}

No analysis here isolates the cause, and we do not attribute the exceedances to selection bias alone. The practical reading is that the point budget was met on TEST and the demonstrated, upper-bound budget was not met everywhere. The PASS verdict rests on the point budget, as pre-registered.

\begin{table}[t]
\centering\small
\caption{TEST operating points selected by each rule {[}S{]}. Bold marks point losses above 1 pp.~The macro savings under the point-only rule are means of the per-benchmark entries.}
\resizebox{\ifdim\width>\linewidth\linewidth\else\width\fi}{!}{%
\begin{tabular}{@{}llllll@{}}
\toprule
Rule & Method & OpenBookQA saving / loss (pp) & CommonsenseQA saving / loss (pp) & HellaSwag saving / loss (pp) & Macro saving \\
\midrule
Point-only & π1-x & 0.305 / \textbf{1.13} & 0.357 / \textbf{1.10} & 0.351 / \textbf{1.07} & 0.337 \\
Point-only & native-pb & 0.403 / \textbf{1.57} & 0.359 / \textbf{1.20} & 0.316 / 0.67 & 0.359 \\
Bootstrap-screened & π1-x & 0.157 / 0.17 & 0.282 / 0.60 & 0.279 / 0.40 & 0.239 \\
Bootstrap-screened & native-pb & 0.136 / 0.20 & 0.161 / −0.10 & 0.262 / 0.13 & 0.186 \\
\bottomrule
\end{tabular}
}
\end{table}

Under the point-only rule on TEST, π1-x exceeds 1 pp on all three benchmarks (1.07 to 1.13 pp) and native-pb on two (OpenBookQA 1.57 pp and CommonsenseQA 1.20 pp) {[}S{]}.

\subsection{Rule-dependent preference}

Which method is preferred depends on which selection rule is applied, and the two rules select different operating points. Under the point-only rule, the selected operating points broke the 1 pp budget on TEST in 5 of the 6 benchmark-method cells (Table 5). We do not claim that a fixed pair of policies reverses its order. Table 5 shows what changes. Moving from the point-only rule to the bootstrap-screened rule, each method retreats to a lower-saving operating point, and native-pb retreats further: its macro saving falls from 0.359 to 0.186, a drop of 0.173, against a drop from 0.337 to 0.239, or 0.098, for π1-x {[}P, differences of the Table 5 macro entries{]}. The largest single change is native-pb on OpenBookQA, which moves from 0.403 at 1.57 pp loss to 0.136 at 0.20 pp loss. The point-only preference for native-pb comes from one benchmark: per benchmark, π1-x − native-pb is −0.098 on OpenBookQA, −0.002 on CommonsenseQA and +0.035 on HellaSwag {[}P, differences of the rounded Table 5 entries; the unrounded CommonsenseQA difference is $-0.00275${]}. On OpenBookQA, native-pb's point-only selection used a threshold of 1.9\(\times 10^{-5}\) on the CDF with depth cap 4, so the statistic barely filtered, and its loss rose from 0.98 pp at selection to 1.57 pp on TEST, the largest overshoot of any selected operating point {[}P{]}. That point-only selections would exceed the budget on new data was visible before TEST: in the held-out folds of E1, point-only selections exceeded 1 pp in 6 of 15 benchmark-by-fold cells for native-pb and 7 of 15 for π1-x, against 3 of 15 each under the bootstrap-screened rule {[}E{]}. The direction of the rule dependence was also consistent across development folds: in each of the five E1 held-out folds, π1-x − native-pb was negative under the point-only rule and positive under the screened rule, with a screened-minus-point-only difference of +0.051 to +0.127 {[}E{]}. E1 motivated C1, so this corroborates rather than independently confirms the TEST pattern, and no interval for the difference between rules was computed on TEST.

The mechanism is not established by our data. One candidate explanation, which we flag as {[}P{]} and as untested, is that the native CDF is a skewed statistic with little resolution near the point where the screen stops accepting thresholds, whereas π1-x's probability is graded more finely there. The only supporting evidence is exploratory: on development data the out-of-fold AUROC of π1-x's score for the stability label exceeded that of the native CDF (0.689 against 0.603), and in the E1 report the point-only thresholds selected for the native CDF were extremely small for some benchmarks {[}E{]}. Neither observation shows that a better ranking is what produces the larger saving under the screened rule; E1 itself records that the ranking quality does not explain the point-only win of the native gate. Neither rule compares the methods at matched loss, and no matched-loss comparison was pre-registered.

\begin{figure}[t]
\centering
\includegraphics[width=\linewidth]{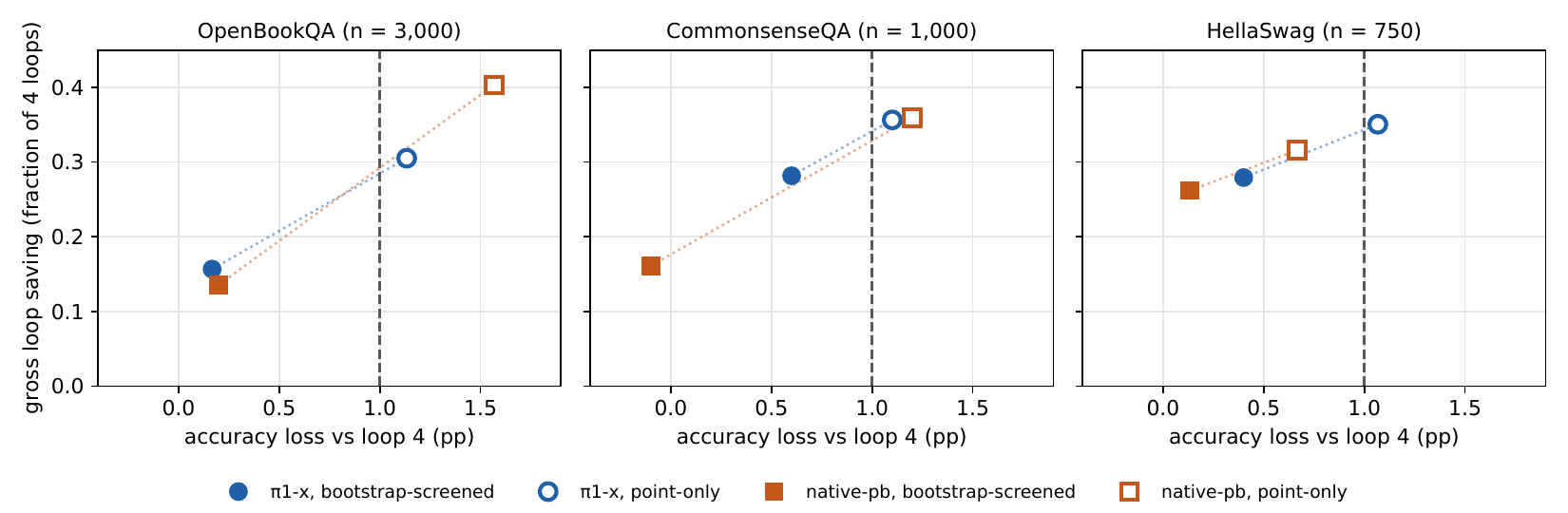}
\caption{TEST operating points selected by each rule, one panel per primary benchmark (OpenBookQA, CommonsenseQA, HellaSwag). Horizontal axis: accuracy loss relative to loop 4, in percentage points. Vertical axis: gross loop saving, as a fraction of the four-loop budget. Circles: π1-x. Squares: native-pb. Filled markers: bootstrap-screened rule. Open markers: point-only rule. The dashed vertical line marks the 1 pp budget. Each marker is one frozen \((\tau_b, m_b)\) selected on development data; the figure shows four operating points per panel and is not a saving-loss frontier, so it does not support a comparison at matched loss. Macro-averaged over the three benchmarks, the screened-rule difference π1-x − native-pb is +0.053 {[}+0.049, +0.057{]} {[}C{]} and the point-only difference is −0.022 {[}−0.026, −0.018{]} {[}S{]}.}
\end{figure}

\subsection{Net saving}

Net of the estimated probe overhead, π1-x saves 0.038 more than native-pb, 95\% CI {[}+0.033, +0.042{]} {[}S, estimated cost{]}. The per-benchmark net savings of π1-x are 0.125 (OpenBookQA), 0.274 (CommonsenseQA) and 0.272 (HellaSwag), against gross savings of 0.157, 0.282 and 0.279; the overhead costs 0.031 on OpenBookQA, where loops are short and the overhead relative to one loop is largest, and 0.007 on each of the other two (differences of the stored values) {[}S{]}. The overhead reduces the advantage but does not remove it. This is an accounting estimate, not a wall-clock measurement, and native-pb is credited with zero controller cost by definition (section 3.5).

\section{Results: RQ2 and RQ3}

\subsection{RQ2: against the native gate and against untrained baselines}

RQ2 asks whether an added trained probe beats (a) Ouro's trained native gate and (b) a training-free exit based on a divergence between successive loops. For (a) the primary comparison in section 5.1 is the answer {[}C{]}: under the bootstrap-screened rule, π1-x saved more than native-pb. Because π1-x takes the native CDF as one of its inputs (Section 3.3), this compares the native statistic with a model that contains it; comparison (b) is the more informative test of what the trained probe adds. In the exploratory E1 analysis, a probe without the CDF feature still saved more than native-pb under the screened rule in all five held-out folds (+0.021 macro over the three primary benchmarks), but not over all six development benchmarks (−0.010) {[}E{]}; it was not evaluated on TEST. For (b), and for every other baseline, the comparisons were pre-specified as secondary {[}S{]}. The emphasis we place on the comparison with r-kl is post hoc {[}P{]}, because r-kl was identified as the strongest baseline using TEST outcomes (below).

\begin{table}[t]
\centering\small
\caption{TEST comparison of π1-x (macro saving 0.239) with every baseline, paired and benchmark-stratified bootstrap, 10,000 resamples {[}S{]}. The r-kl row is given to four decimals. The nine intervals are not adjusted for multiple comparisons. All baselines were selected on development data under the bootstrap-screened rule.}
\resizebox{\ifdim\width>\linewidth\linewidth\else\width\fi}{!}{%
\begin{tabular}{@{}llll@{}}
\toprule
Baseline (selected under the same rule) & Macro gross saving & Δ = π1-x − baseline & 95\% CI \\
\midrule
B\_pb & 0.083 & +0.156 & {[}+0.151, +0.161{]} \\
native-pb (primary) & 0.186 & +0.053 & {[}+0.049, +0.057{]} \\
r-conf & 0.171 & +0.068 & {[}+0.063, +0.073{]} \\
r-margin & 0.170 & +0.069 & {[}+0.064, +0.075{]} \\
r-ent & 0.179 & +0.060 & {[}+0.055, +0.065{]} \\
r-dlogit & 0.162 & +0.077 & {[}+0.071, +0.083{]} \\
r-hcos & 0.174 & +0.065 & {[}+0.059, +0.072{]} \\
r-hl2 & 0.149 & +0.090 & {[}+0.084, +0.096{]} \\
\textbf{r-kl} & \textbf{0.238} & \textbf{+0.00125} & \textbf{{[}−0.0041, +0.0065{]}} \\
\bottomrule
\end{tabular}
}
\end{table}

All the readouts other than r-kl trail π1-x by between 0.060 and 0.090, and fixed depth trails by 0.156 {[}S{]}. Every baseline's TEST point loss stayed within 1 pp (the largest was 0.93 pp, for r-dlogit on HellaSwag) {[}S{]}; upper bounds were computed only for π1-x, native-pb and B\_pb.

\textbf{r-kl.} The difference between π1-x and r-kl is +0.00125, with a 95\% interval of {[}−0.0041, +0.0065{]} {[}S{]}. The strongest baseline was determined on TEST: in a bootstrap that re-selected the strongest baseline in every resample, r-kl was selected in 10,000 of 10,000 resamples, and the re-selected interval is the same as r-kl's own. Because the comparator was chosen with TEST outcomes, this comparison was always secondary and never touched the verdict. The result is non-superiority: we did not demonstrate that the added trained probe is better than r-kl. It is not a demonstration that the two perform alike. No margin for declaring them interchangeable was pre-specified or tested. The interval excludes only differences larger than 0.0065 in favour of π1-x or 0.0041 in favour of r-kl, in units of the four-loop budget {[}S{]}. The prominence we give this comparison is post hoc {[}P{]}.

Development data gave mixed signals about this near-tie. On the development data used for selection, the in-sample macro savings were 0.246 for π1-x and 0.243 for r-kl, against 0.193 for native-pb {[}E, in-sample, not a held-out estimate{]}. The nested, cross-fitted E1 estimate under the same rule disagreed: there, π1-x reached 0.258 and r-kl only 0.207, against 0.204 for native-pb {[}E{]}. The TEST result matches the in-sample freeze numbers rather than the held-out E1 estimate. One candidate reason is that E1's nested estimate refitted the probe and reselected thresholds in each fold, whereas the freeze and TEST used one fit on all development data. This is a {[}P{]} conjecture that we have not tested. On TEST, r-kl exceeded native-pb by 0.052 in macro saving (0.238 against 0.186; a difference of the stored macro values, with no interval computed) {[}P{]}. The confirmatory claim concerns π1-x only; the comparison of r-kl with native-pb is descriptive. The evidence does not show that the learned probe adds value over a simple stability statistic.

\textbf{Relation to the KL exit of Geiping et al.} The rule r-kl is close in spirit to the convergence exit of \citet[section 6.1]{Geiping2025}, which stops a recurrent-depth model when the KL divergence between successive steps falls below 5\(\times 10^{-4}\), once per generated token and with a fixed threshold. Our r-kl differs in the unit and in the threshold. The unit is the question, and the distribution compared is the distribution over the answer options, not over the vocabulary for a generated token. The threshold is not fixed: for each benchmark, a threshold and a cap are selected on development data under the screened rule. The frozen thresholds correspond to halting when \(\mathrm{KL}(p_t \| p_{t-1})\) falls to or below 3.5\(\times 10^{-3}\) on OpenBookQA (cap 3), 3.7\(\times 10^{-2}\) on CommonsenseQA (cap 4) and 2.7\(\times 10^{-4}\) on HellaSwag (cap 3) {[}S, frozen at freeze{]}, so only one of the three is of the order of the Geiping et al.~value. We therefore do not describe r-kl as a reproduction of that exit, and our comparison does not test it.

\subsection{RQ3: out-of-distribution budget}

RQ3 asks whether development-selected policies keep their accuracy budget on unfamiliar benchmarks. For PIQA and Social IQa, each policy used one task-agnostic pair \((\tau, m)\), selected on the development data under the same bootstrap-screened rule, qualifying only if every one of the six development benchmarks met the rule: π1-x with \(m = 4\) and \(\tau = 0.961\), native-pb with \(m = 4\) and \(\tau = 0.386\) (development macro saving 0.159 for each). Both benchmarks are evaluated 0-shot, which differs from the few-shot prompts of most development benchmarks and is part of the shift. The protocol listed this analysis as a secondary one with a verdict-style point-estimate table per benchmark, and never pooled across benchmarks {[}S{]}.

\begin{table}[t]
\centering\small
\caption{Out-of-distribution TEST results with one task-agnostic \((\tau, m)\) per policy {[}S{]}. Bold marks losses and upper bounds above 1 pp.~The B\_pb row is full depth.}
\resizebox{\ifdim\width>\linewidth\linewidth\else\width\fi}{!}{%
\begin{tabular}{@{}llllll@{}}
\toprule
Benchmark (n) & Method & Gross saving & Loss (pp) & Upper bound (pp) & Exits at loop 1 / 2 / 3 / 4 \\
\midrule
PIQA (1,837) & π1-x & 0.308 & 0.71 & \textbf{1.52} & 0 / 433 / 1,399 / 5 \\
PIQA & native-pb & 0.126 & 0.11 & 0.60 & 0 / 0 / 923 / 914 \\
PIQA & B\_pb & 0.000 & 0.00 & 0.00 & 0 / 0 / 0 / 1,837 \\
Social IQa (1,930) & π1-x & 0.481 & \textbf{3.26} & \textbf{4.72} & 0 / 1,785 / 143 / 2 \\
Social IQa & native-pb & 0.028 & −0.05 & 0.16 & 0 / 0 / 214 / 1,716 \\
Social IQa & B\_pb & 0.000 & 0.00 & 0.00 & 0 / 0 / 0 / 1,930 \\
\bottomrule
\end{tabular}
}
\end{table}

On PIQA, π1-x met the point budget (0.71 pp) but its upper bound (1.52 pp) was above 1 pp, and it saved 0.183 more than native-pb {[}S{]} (95\% CI {[}+0.174, +0.191{]}). On Social IQa, π1-x did not stay within the budget: it lost 3.26 pp, with an upper bound of 4.72 pp, and it halted at loop 2 for 1,785 of 1,930 questions. Its large saving there (0.481) comes from halting early on most questions, and the accuracy cost is real: accuracy fell from 43.32\% at loop 4 to 40.05\% {[}S{]}. The saving difference against native-pb, +0.453 (95\% CI {[}+0.449, +0.458{]}), should not be read as an advantage, because it is paid for by the budget violation. The native gate stayed within the budget on both benchmarks, at a much smaller saving (0.126 and 0.028).

We draw no general conclusion from this. It is one failure on one benchmark, with development-selected, task-agnostic thresholds, on two benchmarks that were also 0-shot and differ from the development data in several ways at once. The repository records a bf16 numerical difference between the packed and context-only passes on the 0-shot sets, and a documented PIQA numerical deviation (section 4.4), but no analysis here links either to the Social IQa result, and we do not offer an explanation. The finding is a reason to report an out-of-distribution check next to any development-selected halting policy, and it is not a statement about out-of-distribution robustness in general {[}S{]}.

\section{Discussion}

\subsection{What the results support}

The evidence supports three statements, each at a different strength.

\begin{enumerate}
\def\labelenumi{\arabic{enumi}.}
\tightlist
\item
  \textbf{Under the bootstrap-screened rule, π1-x saved more loops than Ouro's native gate on the three primary benchmarks} {[}C{]}. The difference
  was +0.053 in gross macro saving, 95\% CI {[}+0.049, +0.057{]} (test-sampling uncertainty for the frozen policies), and +0.038 after the estimated probe overhead {[}S{]}. The selected points differ in loss (macro 0.39 against 0.08 pp on TEST), so this is not a comparison at equal accuracy.
\item
  \textbf{Under two pre-specified selection rules, on this model, the preferred method differed} {[}S{]}, contrasted with the {[}C{]} result. Under the point-only rule, the native gate was preferred on development data (G3a, −0.024 {[}E{]})
  and on TEST (−0.022 {[}−0.026, −0.018{]} {[}S{]}). Under the bootstrap-screened rule, π1-x was preferred {[}C{]}. Each rule selects its own
  operating points, so this is a statement about comparisons made through a selection procedure, not about fixed policies changing order.
  The point-only preference for the native gate on TEST compares operating points that broke the observed 1 pp budget on TEST: native-pb on two of
  three benchmarks, π1-x on all three (Table 5). That comparison therefore shows which method a point-only rule prefers. It does not show which method
  is better among budget-compliant policies. On development data, the same direction appeared in each of the five E1 held-out folds {[}E{]}.
\item
  \textbf{The added trained probe showed no demonstrated advantage over an untrained KL-between-loops readout} {[}S{]}, and this comparison is
  emphasised post hoc {[}P{]}. Separately, one out-of-distribution benchmark showed a budget failure for π1-x {[}S{]}.
\end{enumerate}

What the evidence does not support:

\begin{itemize}
\tightlist
\item
  a claim that π1-x is a better halting signal than prior art;
\item
  a claim that π1-x and r-kl are equivalent;
\item
  any statement about latency or about Ouro's released inference path;
\item
  any statement about models, loop counts or task formats other than those tested;
\item
  any general statement about robustness to distribution shift.
\end{itemize}

\subsection{Relation to prior work}

\textbf{Post-hoc readouts on Ouro.} \citet{Popescu2026} already reported that post-hoc trajectory readouts can match or beat Ouro's
pretrained gate at validation-selected operating points. They describe the gate as ``competitive but not uniformly Pareto-optimal'' and
conclude that adaptive depth is ``a trajectory--readout problem, not a gate-learning problem''. Our primary result is consistent with that view
and should be read as a pre-registered, sealed-test instance of it under one selection rule, not as a new finding that readouts can beat the
gate. What we add is the following:

\begin{itemize}
\tightlist
\item
  the preference depends on the selection rule (a point-only rule favours the native gate on our data). Popescu et al.~report one
  validation-selected operating point per method in their tables and do not state the selection criterion. They do not compare selection
  rules, accuracy-loss budgets or statistical bounds. Their full accuracy-versus-loops curves on Ouro (their Figures 19--20) interleave and
  cross on several benchmarks, so a dependence of the preferred method on the operating point is visible in their data, although they do not
  discuss it. What we add is narrower: a demonstration that two pre-registered accuracy-budget rules lead to opposite preferences on a sealed
  test set, and that the stricter rule's bounds do not all hold there;
\item
  the readout that matches the trained probe is a divergence between loops rather than a confidence readout {[}P{]}. r-kl was identified as the strongest baseline using TEST outcomes. Popescu et al.~list hidden-state cosine similarity and displacement among their more promising readouts on Ouro; in our per-question protocol neither stood out (r-hcos 0.174 and r-hl2 0.149 macro, the latter the lowest of any readout, against 0.238 for r-kl; Table 6);
\item
  a development-selected policy can fail its budget on an out-of-distribution benchmark.
\end{itemize}

\textbf{Risk control for early exit.} Fast yet Safe \citep{Jazbec2024} selects early-exit thresholds with a formal guarantee (a UCB bound at level 1−δ under
i.i.d. calibration data). Learn then Test \citep{Angelopoulos2025LTT} and Conformal Risk Control \citep{Angelopoulos2024CRC} give the general machinery. Our
selection rule is not one of these procedures and inherits none of their guarantees (Section 3.4). The TEST upper bounds that exceeded 1 pp
(Section 5.3) are compatible with a heuristic screen, with small test samples and with development--test differences {[}P{]}. They argue for a
formal procedure on separate calibration data, not for treating our screen as one. A natural follow-up is to repeat the comparison with LTT or UCB selection, on calibration data separate from the data used for fitting.

\textbf{Calibration and ranking quality.} \citet{Kubaty2026} show that calibration alone is insufficient to measure the quality of early-exit classifiers. We see an analogous gap
for ranking quality in our exploratory data. Better ranking of the stability label (π1-x AUROC 0.689 against 0.603 for the native CDF {[}E{]}) did not stop
the native gate from winning under the point-only rule {[}E{]}, so ranking quality alone does not predict which method a given selection rule will prefer.

\textbf{Convergence exits.} Our r-kl is close in spirit to the KL convergence exit of \citet{Geiping2025}. It differs in unit (the question,
over answer options), in the distribution compared, and in threshold choice (selected per benchmark, not fixed). We therefore claim no
reproduction of their exit. In our setting we found no demonstrated advantage of the trained probe over a convergence statistic {[}S{]}, {[}P{]}.

\subsection{Why the preference might depend on the rule}

We have no causal analysis. The following are hypotheses for future work {[}P{]}:

\begin{itemize}
\tightlist
\item
  \textbf{Resolution near the boundary.} The two rules accept different regions of the threshold grid. The point-only rule accepts riskier operating
  points, and there the native CDF reached higher savings (macro 0.359 against 0.337). Adding the upper-bound screen removed those points for both
  methods, and the native gate lost more (−0.173 against −0.098). A statistic with coarse resolution near the acceptance boundary would behave this
  way. We did not test this.
\item
  \textbf{Correctness flips and the bound.} A property of the screen, not a hypothesis: the per-question loss difference takes the values −1, 0 and +1, so the bootstrap bound widens with the number of questions whose correctness changes in either direction. Unlike the point-only rule, the screened rule therefore penalises gains and losses that cancel. A hypothesis: π1-x, trained to predict that the answer will not change again, may reach a given saving with fewer such flips, which would explain part of its advantage under the screened rule. On development data, in-sample, the selected π1-x points had smaller bound margins than native-pb's on all three benchmarks while saving more (OpenBookQA 0.44 against 0.53 pp, CommonsenseQA 0.51 against 0.68 pp, HellaSwag 0.37 against 0.38 pp); this partly restates the selection and is not independent evidence. On TEST the margins were mixed (Table 4). We examined this hypothesis on development data in a protocol-frozen exploratory analysis (X1; Section 4.4). At matched gross saving on held-out folds, π1-x changed the correctness of fewer questions than native-pb, in both directions, at each of the four pre-declared saving levels (for example 1.3 against 3.3 flips per 100 questions at saving 0.20), with no demonstrated difference in net loss {[}E{]}. This satisfies X1's pre-declared exploratory decision rule; the four levels are related and the intervals are unadjusted, so it is not four independent confirmations. The bootstrap intervals condition on the matched thresholds and the fitted probes and do not capture uncertainty from fitting or matching. The result is consistent with the screen's flip penalty contributing to π1-x's advantage under the screened rule; it is not a causal demonstration. Three qualifications apply. The reduction was concentrated on CommonsenseQA. It does not explain r-kl: the untrained KL readout produced more flips than native-pb at matched saving, yet matched π1-x under the screened rule on TEST {[}E{]}. And X1 compares methods at matched gross saving, not matched accuracy loss, so it does not show a better accuracy trade-off.
\item
  \textbf{Selection noise.} Both rules select one of 51 thresholds per benchmark on the same development data used to judge them. Part of any
  difference between rules may be selection noise rather than a property of the signals. The disagreement between the in-sample freeze
  numbers and the cross-fitted E1 estimate for r-kl (Section 6.1) shows that this noise is not negligible.
\end{itemize}

\subsection{Implications for evaluating halting methods}

\begin{enumerate}
\def\labelenumi{\arabic{enumi}.}
\tightlist
\item
  Treat a halting comparison as a comparison of signal plus selection procedure. Report results under more than one operating rule, and report
  the selected operating points, not only the macro saving.
\item
  Report upper bounds on TEST alongside point losses. Under the primary bootstrap-screened selection rule, all nine reported benchmark-by-policy point losses were within the 1 pp budget, but four of their nine corresponding upper bounds exceeded it.
\item
  Where a guarantee is wanted, use a formal procedure (LTT, CRC, UCB as in Fast yet Safe) with held-out calibration data, and state its assumptions.
\item
  Include strong untrained baselines before attributing gains to a trained controller. In our study the strongest untrained baseline was a divergence between successive loops.
\item
  Include at least one out-of-distribution check of a development-selected policy. A policy can save the most precisely by halting where it should not.
\item
  Keep simulated loop savings apart from measured latency, and state whether policies were executed or evaluated from stored traces.
\end{enumerate}

\subsection{Limitations}

\begin{itemize}
\tightlist
\item
  \textbf{One model, one size, four loops.} Every result is for Ouro-1.4B with 4 loops in bf16. Ouro-2.6B, more loops and other looped architectures were not tested.
\item
  \textbf{Multiple-choice scoring, not generation.} Each question has one decision, at the last context token, over a packed sequence with all
  options. Per-token halting during generation, the setting of Ouro's released code and of Geiping et al., is a different problem.
\item
  \textbf{Our decision unit is an adaptation.} Ouro's released code computes every loop and selects a readout for each token position. Our
  per-question stopping, with the skipped loops counted as saved, is not Ouro's mechanism, and our savings are not measurements of Ouro's
  released inference.
\item
  \textbf{The native-gate baseline is used off its training objective.} Ouro's gate was trained with the model for per-token language modelling, and we use it as a per-question statistic at one position. π1-x also takes the native CDF as one of its inputs (Section 3.3), so the primary comparison is against a statistic the probe contains. The comparison with the untrained r-kl (Section 6.1) is the more informative one. An exploratory probe without the CDF feature still led native-pb under the screened rule on the three primary development benchmarks, but not over all six (Section 6.1) {[}E{]}.
\item
  \textbf{Retrospective evaluation and simulated savings.} Policies were applied to stored full-depth traces. Exactness under truncation was checked
  by an executed truncated run on 600 development questions (D-t; all 1,800 policy-question decisions agreed), not on the test set, and the per-loop statistics were not bitwise reproducible across GPU runs (Section 3.6). Savings are loop counts plus an overhead measured separately at freeze time.
  Apart from the descriptive D-t timings on development data, no held-out end-to-end latency evaluation of the halting policies was performed. A loop is assumed to cost the same whether or not later loops follow, and the native gate is assigned zero
  controller cost.
\item
  \textbf{Numerical path.} The planned check of whether halting decisions agree across bf16 numerical paths (Phase 3 check D-b) never ran, because
  Phase 3 stopped at G3a. A bf16 difference between the packed and context-only passes is documented on the 0-shot sets. Its effect on halting
  decisions is unmeasured.
\item
  \textbf{No formal guarantee.} The bootstrap screen is asymptotic, uncorrected for the 51 candidates, and computed on the data that selected the
  candidate.
\item
  \textbf{Benchmark identity is used.} The primary policies select \((\tau_b, m_b)\) per benchmark, so a deployment would need to know the task. The
  out-of-distribution analysis used one task-agnostic pair and is the closer model of deployment. That is where π1-x failed.
\item
  \textbf{Sample size.} HellaSwag has \(n = 750\), which limits the precision of its upper bound. Every method, including fixed depth, exceeded 1 pp there.
\item
  \textbf{Out-of-distribution scope.} There are two benchmarks, both 0-shot, which differ from the development data in format and in task at once,
  so the source of the Social IQa failure cannot be isolated.
\item
  \textbf{The TEST set is spent.} No further confirmatory analysis can use it. Any claim about r-kl against the native gate, or about an
  out-of-distribution guard, needs new held-out data.
\end{itemize}

\subsection{Threats to validity}

\begin{itemize}
\tightlist
\item
  \textbf{Internal.}

  \begin{itemize}
  \tightlist
  \item
    r-kl was singled out using TEST outcomes. Its comparison with π1-x was pre-specified, but its prominence is not.
  \item
    The in-sample freeze numbers and the cross-fitted E1 estimate disagree about r-kl.
  \item
    The Phase 3 float32 check on PIQA (D-a) failed its pre-set tolerance for the difference between packed and context-only scoring. A
    follow-up with a float64 reference (D-a2; maximum float64 difference 1.67\(\times 10^{-13}\)) classified the difference as numerical. The reviewer
    accepted this as a documented exception, not a retroactive pass.
  \item
    The console transcript of the TEST run was overwritten. The run's outputs are fully recorded in the report and the marker.
  \item
    The bootstrap-screened rule was adopted for C1 after the exploratory E1 found that π1-x led under it on development data. The sealed test shows that this development finding held on new data; it does not show that the rule was chosen without knowledge of development outcomes.
  \end{itemize}
\item
  \textbf{Construct.}

  \begin{itemize}
  \tightlist
  \item
    Gross saving counts loops, not FLOPs. Net saving uses ρ timed in isolation and is an accounting estimate.
  \item
    The 1 pp budget is relative to loop-4 accuracy on the same questions, not to an external standard.
  \end{itemize}
\item
  \textbf{Statistical.}

  \begin{itemize}
  \tightlist
  \item
    The primary interval is a paired, benchmark-stratified percentile bootstrap. It does not reflect variability from refitting the probe or
    reselecting thresholds, so it conditions on the frozen policies.
  \end{itemize}
\item
  \textbf{External.}

  \begin{itemize}
  \tightlist
  \item
    The results cover one model family, multiple-choice scoring and five benchmarks.
  \end{itemize}
\end{itemize}

\section{Conclusion}

On a frozen Ouro-1.4B, a pre-registered comparison on a sealed test set found that a trained stability probe saved more loops than Ouro's
native gate when operating points were chosen with a bootstrap-screened accuracy rule: +0.053 gross macro saving {[}C{]}, +0.038 net (estimated) {[}S{]}.
The same comparison under a point-only rule preferred the native gate, both on development data {[}E{]} and on the sealed test set {[}S{]}. On the test set,
however, the point-only operating points of both methods exceeded the observed 1 pp budget on most benchmarks, so the reversal concerns which method each
rule prefers, not which method is better among compliant policies.

The trained probe showed no demonstrated advantage over an untrained KL-between-loops readout. That comparison was pre-specified {[}S{]}, and we
give it post hoc emphasis {[}P{]}. On one out-of-distribution benchmark the probe broke its accuracy budget {[}S{]}.

These results are consistent with prior evidence that simple trajectory readouts compete with Ouro's learned gate \citep{Popescu2026}. Their main
lesson is methodological: which halting method looks better depends on how operating points are selected. Halting methods should be compared
under explicit selection procedures, preferably ones with formal risk control, alongside strong untrained baselines and out-of-distribution
checks.

Future work needs new held-out data:

\begin{itemize}
\tightlist
\item
  a pre-registered comparison of KL exit and the native gate;
\item
  formal risk-controlled selection;
\item
  an out-of-distribution guard;
\item
  Ouro-2.6B;
\item
  a truncated implementation with measured end-to-end time on held-out data.
\end{itemize}

\section*{Reproducibility and data availability}
The code, the pre-registered and exploratory protocols, the decision log, the result artifacts and the audit reports are in the project repository (\codeavailability). A script recomputes the reported numbers from the committed result artifacts on a CPU. The development-stage computations (the frozen selections, the held-out diagnostics and the correctness-flip analysis) were recomputed from the stored development features, bit for bit, in the original software environment, and to numerical agreement on a second platform.

Some components cannot be reproduced from the repository. The extracted per-question features are not released; regenerating them needs a CUDA GPU and the pinned model revision, and GPU extraction is not bitwise reproducible (Section~3.6). The sealed test features are not released and the test set has been used once, so the test statistics and their bootstrap intervals are verified against pinned file hashes and the recorded hash of the code that produced them, not independently recomputed. The benchmarks are public and are not redistributed here; the manifests contain identifiers only.

\section*{Statement on AI assistance}
AI-assisted tools were used for code development, test generation, manuscript drafting and editing, and reproducibility auditing. AI-assisted methodological review was also used to identify potential weaknesses and refine the interpretation and presentation of results.

The author designed the study, directed the experiments, reviewed the analyses, and made the scientific decisions documented in the project's decision log. The author takes responsibility for the methodology, results, interpretations, and final manuscript. AI-assisted review does not constitute independent human peer review.